\documentclass[lettersize, journal]{IEEEtran}

\IEEEoverridecommandlockouts                              

\usepackage{amssymb}  
\usepackage{graphicx} 
\usepackage{float} 
\usepackage{subfigure} 
\usepackage{amsmath}
\usepackage{mathtools} 
\usepackage{booktabs} 
\usepackage{array}    
\usepackage{multirow} 
\usepackage{caption}
\usepackage{algorithm}
\usepackage{algpseudocode}
\usepackage{cite}
\usepackage{url}
\usepackage[table]{xcolor}
\usepackage{siunitx}
\usepackage{threeparttable}
\usepackage{adjustbox}

\title{\LARGE \bf
V2I-Calib++: An Online Spatial Calibration Approach for Multi-end LiDAR in Urban Intersections
}

\author{Qianxin Qu$^{\ast}$, Xinyu Zhang$^{\ast}$, Yifan Cheng$^{\ast}$, Yijin Xiong$^{\ast\dagger}$, Chen Xia, Ziqiang Song, Qian Peng, Kang Liu, Jun Li

\thanks{This research has been supported by the National Natural Science Foundation of China (Project No. 52221005; 62273198; 52204180), and the Beijing Natural Science Foundation Program under Grant No.L241017, and the National Research Foundation Singapore under its AI Singapore Programme (Award Number: AISG2-GC-2023-007) and the Tsinghua University-Didi Future Mobility Joint Research Center.}

\thanks{ the State Key Laboratory of Automotive Safety and Energy, and the School of Vehicle and Mobility, Tsinghua University, Beijing, 100084, China.}
\thanks{$^{\ast}$ These authors contributed equally to this work as co-first authors}
\thanks{$^{\dagger}$ Corresponding author.}
}

\begin{document}

\maketitle
\thispagestyle{empty}
\pagestyle{empty}


\begin{abstract}

Urban intersections, dense with pedestrian and vehicular traffic and compounded by positioning signal obstructions, are among the most challenging areas in urban traffic systems. Traditional single-vehicle intelligence systems often perform poorly in such environments due to a lack of global traffic flow information and the ability to respond to unexpected events. Vehicle-to-Everything (V2X) technology, through real-time communication between vehicles (V2V) and vehicles to infrastructure (V2I), offers a robust solution. However, practical applications still face numerous challenges. Calibration among vehicle and infrastructure endpoints with different configurations in multi-end LiDAR systems is crucial for ensuring the accuracy and consistency of perception system data. Most existing multi-end calibration methods rely on initial calibration values provided by positioning systems, but the instability of GNSS signals due to high buildings in urban canyons poses severe challenges to these methods. To address this issue, this paper proposes a novel multi-end LiDAR system calibration method that does not require positioning priors to determine initial external parameters and meets real-time requirements. Our method introduces an innovative multi-end perception object association technique that leverages a new \emph{Overall Distance} (\emph{oDist}) metric to measure the spatial association between perception objects, subsequently using this metric as the foundation for an optimal transport formulation. By this means, we can extract co-observed targets from object association results for further external parameter computation and optimization. Extensive comparative and ablation experiments conducted on the simulated dataset V2X-Sim and the real dataset DAIR-V2X confirm the effectiveness and efficiency of our method. The code for this method can be accessed at: \url{https://github.com/MassimoQu/v2i-calib}.

\end{abstract}


\section{INTRODUCTION}

Urban intersections, as key nodes of urban transportation networks, handle significant volumes of pedestrian and vehicle traffic daily. Particularly in city centers and commercial districts, the road junctions at these sites exhibit highly dynamic traffic flows, not only increasing the risk of traffic accidents \cite{lin2021intelligent} but also posing substantial challenges to traditional traffic management systems.

In complex environments, conventional intelligent transportation systems often struggle to adapt due to limited traffic flow information and insufficient emergency response capabilities \cite{huang2021dynamic}. To address these challenges, Vehicle-to-Everything (V2X) technology \cite{luo2021intersection} integrates real-time data exchange between vehicles (V2V) \cite{yang2023dynamic} and infrastructure (V2I) \cite{sun2022eco}, enabling efficient management of traffic scenarios, such as congestion and accidents \cite{lin2021intelligent}, and significantly improving urban traffic safety and efficiency.

\begin{figure}[tbp] 
\centering 
\includegraphics[width=0.5\textwidth]{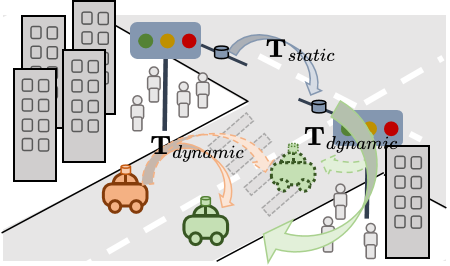} 
\caption{Schematic of calibration challenges at urban intersections.} 
\label{fig:static_dynamic} 
\end{figure}

\begin{figure*}[htbp] 
\centering 
\includegraphics[width=\textwidth]{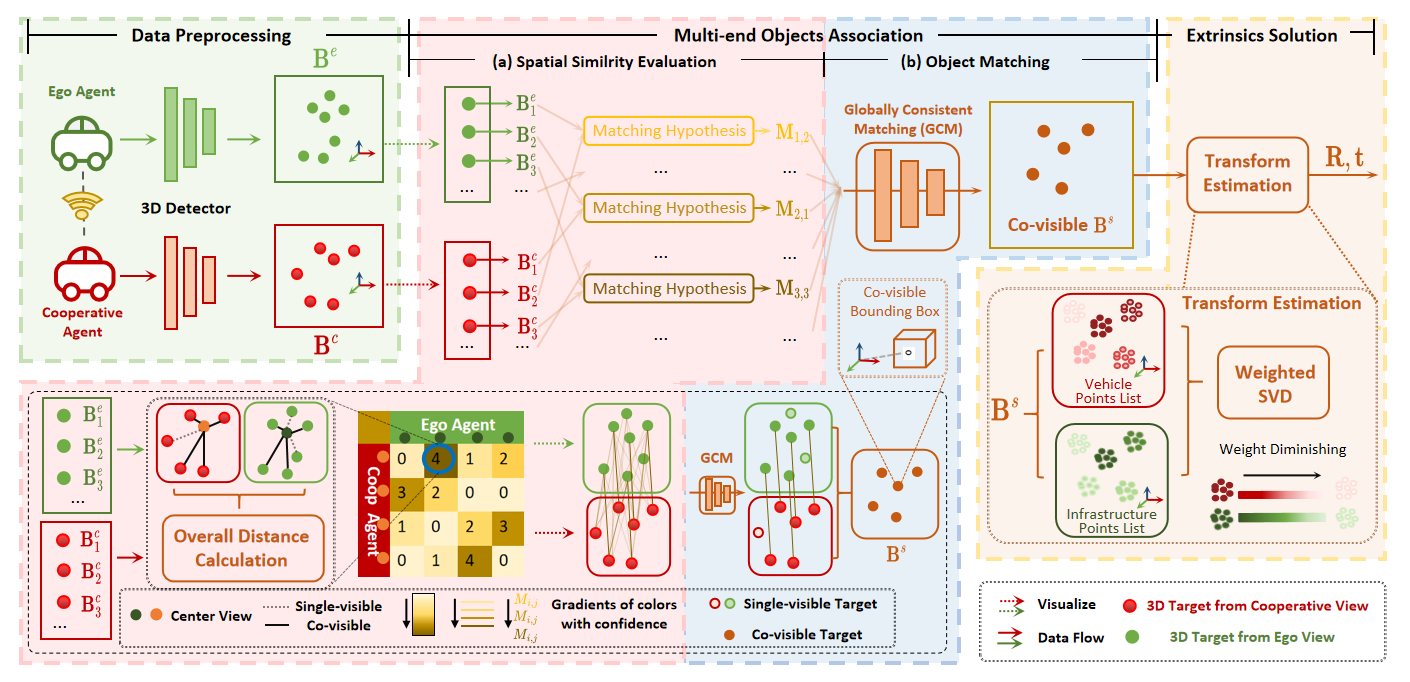} 
\caption{The proposed method first generates 3D detection boxes on each endpoint, then identifies the common objects across endpoints. This multi-end object association is achieved through two steps: spatial similarity evaluation and object matching. The core process involves filtering the association of multi-end 3D detection boxes based on the affinity matrix (evaluated through oDist Calculation, with visualizations of both the matrix and the association degree). After identifying the common objects, feature point clouds are extracted from the detection boxes, and the extrinsic parameters are further computed using weighted point cloud registration.}
\label{fig:V2I-Calib Workflow} 
\end{figure*}

However, this technology also introduces new challenges, particularly in the areas of multi-sensor data fusion and calibration. In V2X systems, different sensors, including multiple LiDAR devices, require precise calibration to ensure data consistency and accuracy. Due to the high dynamism of urban intersection scenarios, traditional static single-time calibration methods \cite{zhang2023automated} no longer meet practical needs. Moreover, existing multi-LiDAR calibration techniques \cite{xu2022v2x,lu2023robust,hu2022where2comm,song2024spatial,VIeyeACM2021, VIPSACM2022 }  generally rely on positioning systems (e.g., GNSS) to provide high-precision initial values, with experimental results \cite{lei2024robust} indicating an error range of 1-2 meters. However, in practice, it is challenging for positioning systems to consistently meet this requirement, limiting the applicability of such methods in real-world scenarios.

In urban environments, the urban canyon effect often causes fluctuations in GNSS signals due to building obstructions and signal reflections. This issue is particularly pronounced at urban intersections, where low vehicle speeds make it difficult to distinguish effective signals from interference \cite{xie2014measuring}, further exacerbating positioning instability. The US Department of Transportation’s V2X project \cite{talas2021system} identified insufficient positioning accuracy at intersections as a major challenge. Related experiments \cite{talas2021cooperative} also revealed that ``missing relative positioning data" is notably more frequent in intersection scenarios, highlighting the difficulty of ensuring reliable positioning accuracy even with advanced systems. In addition, malicious attacks targeting V2X positioning systems pose another significant threat to their stability at urban intersections. These challenges create a fundamental conflict between the positioning demands and inherent instability of urban intersections, making this a critical bottleneck for the widespread adoption of V2X technologies.

To address this issue, this paper proposes an innovative online multi-end LiDAR calibration method—V2X-Calib++. By fully utilizing the perception results from multi-source LiDAR data of vehicles and roadside units, V2X-Calib++ achieves real-time spatial consistency of sensor data without the need for external positioning support, ensuring efficient and reliable operation of V2X systems in complex urban traffic scenarios.

The core of our method is a strategic shift from direct cross-source point cloud processing to leveraging the spatial topology of 3D detected objects. This perception-guided approach dramatically reduces computational complexity and enhances robustness to noise and partial overlaps typical in V2X environments.
The method consists of two main components: a multi-end target association algorithm and a shared target external parameter solution. 
In the multi-end target association component, we introduce a novel \emph{Overall Distance} (\emph{oDist}) metric. This metric is informed by initial SVD-based transformation hypotheses for candidate object pairs and quantifies scene-level spatial consistency. This rich oDist measure then serves as the basis for an Optimal Transport (OT) formulation to achieve robust object matching. Subsequently, the feature point clouds derived from these confidently matched objects are used in a weighted SVD algorithm, where weights reflect matching confidence, to derive the final extrinsic parameters.

While classical point cloud registration techniques often rely on iterative closest point algorithms with SVD for refinement or apply OT to point-level correspondences, these approaches struggle with the high outlier ratios, lack of initial alignment, and scene-level complexity inherent in V2X scenarios \cite{cross-source-pcarxiv2022}. V2X-Calib++ differentiates itself by operating at the semantic level of detected objects, which is inherently suited to tackling the aforementioned V2X challenges. Methodologically, its key innovation is the synergistic framework of SVD and OT: SVD is not merely a final solver but an initial engine for generating object-pair transformation hypotheses that underpin our oDist metric. This oDist then empowers a more robust and meaningful OT stage for global object association, a distinct departure from traditional point-based applications of these techniques.

The main advantages of V2X-Calib++ include its independence from external positioning initial values and its ability to meet real-time requirements. Furthermore, this method leverages traffic participant information commonly present in traffic scenes, enhancing its versatility. By processing information using only perception data from target detection, it elegantly addresses the challenge of high outlier ratios in cross-source point clouds arising from different LiDAR sensor configurations. Compared to other methods requiring complex data processing \cite{VIeyeACM2021}, V2X-Calib++ has lower computational complexity and data transmission costs, making it more suitable for practical applications.

The innovations of this paper are summarized as follows:

\begin{itemize} 
\item [1)] An initial-value-free online calibration method for vehicle-road multi-end scenarios is proposed, based on perception object results, particularly suited for environments like urban canyons where positioning fails; 
\item [2)] A new multi-end target association method is proposed, which fully explores spatial associations in the scene without positioning priors, and its core metric, \emph{Overall Distance} (\emph{oDist}), enables real-time monitoring of external parameters in the scene; 
\item [3)] The effectiveness of the method is validated on both simulated and real datasets, achieving real-time calibration of external parameters. 
\end{itemize}

In the preliminary conference paper \cite{qu2024v2i}, we proposed a framework for initial-value-free LiDAR calibration. This article presents crucial enhancements for improved performance, robustness, and broader applicability. Notably, we have refined our core oDist metric (see Section \ref{sec:odist}), superseding our previous metric, yielding greater computational efficiency (see time comparisons in Table \ref{tab:v2xsim_ideal_compare} and Table \ref{tab:contrast_experiment}) and providing a more robust indicator of scene consistency (referencing the experiment in Fig. \ref{fig:odistance_vs_oIoU}), further substantiated by new real-world application examples (Section \ref{sec:practical_use_cases_v2x_calib_pp}). Additionally, V2I-Calib++ incorporates a novel weighted mechanism into the final extrinsic parameter estimation stage (see Section \ref{sec:extrinsic_solution}), which utilizes association confidences to significantly boost calibration accuracy. The system's capabilities and resilience are also more comprehensively demonstrated through expanded experimentation, including evaluations on a new simulated V2X-Sim dataset (see Table \ref{tab:v2xsim_ideal_compare}) and in-depth comparative analysis (see Table \ref{tab:contrast_experiment}). These developments collectively offer a more mature and rigorously validated solution for initial-value-free LiDAR calibration in complex V2X environments.


\section{RELATED WORK}
\label{section:related work}

This paper focuses on the multi-end LiDAR spatial calibration method in urban intersection scenarios. The core objective of this method is to solve the extrinsic parameters between different LiDARs, i.e., to determine their relative pose relationships, thereby achieving the registration of multi-end point cloud data. The primary goal of point cloud registration is to align the local point clouds captured by each sensor to a common coordinate system. The LiDAR calibration mentioned in this paper refers specifically to spatial calibration, which, in data terms, is a point cloud registration task. We do not distinguish between these two concepts throughout the paper.

Some existing point cloud registration methods can be applied to the scenario described in this paper. We categorize these methods into three groups: Target-based, Targetless, and Deep Learning-based (DL-based) methods.

\subsection{Target-Based Methods} 
Target-based methods typically involve placing calibration targets or boards with known dimensions and geometric features in the scene \cite{dhall2017lidar}, and achieving high-precision extrinsic parameter estimation by detecting these corresponding features. The advantage of such methods lies in the straightforward extraction and matching of local features for registration, as well as controllable calibration accuracy. However, in large-scale dynamic settings such as urban intersections, calibration targets may be occluded or cannot be left in place for an extended period. This inherent limitation underscores their offline nature, making them unsuitable for the demands of online calibration and impractical for real-world deployment.

\subsection{Targetless Methods} 
Targetless approaches do not rely on additional calibration targets and can be further divided into motion-based and scene-based methods:

\textbf{Motion-based methods} \cite{lv2020targetless} utilize the trajectories of the sensor carrier or moving objects in the scene, together with geometric constraints, to estimate relative poses among sensors. These methods generally perform well in scenarios with stable and observable motion. However, if the scene is highly dynamic or if the motion trajectory is not controllable, their accuracy and robustness can degrade significantly.

\textbf{Scene-based methods} extract environmental features (e.g., edges, planes) or apply ICP-style iterative closest point algorithms \cite{pc-registrationarxiv2021, yang2015go, VGICPICRA2021} to register multi-sensor point clouds. These approaches do not require manual placement of calibration objects. Nevertheless, when initial values is inaccurate or in highly dynamic environments, these algorithms may fail to register point clouds or may suffer from reduced accuracy \cite{zheng2022global,shi2021keypoint}. To address these challenges , some researchers have introduced global point cloud registration \cite{yang2020teaser, qin2023geotransformer, zhou2016fast} or graph-based optimization frameworks \cite{gao2021regularized, zhao2024graph, li2023graphps, yang2023global} to reduce dependence on initial values and overlap areas. However, most of these algorithms still demand considerable computational resources for real-time performance.

\subsection{Deep Learning (DL)-Based Methods} 
In recent years, deep learning has made remarkable advances in the extraction and matching of 3D point cloud features, enabling end-to-end calibration/registration \cite{PointnetlkCVPR2019, wang2019deep}. Compared with classical geometric methods, deep learning can learn feature representations that are more robust to noise and dynamic interference using large-scale data. Nonetheless, these approaches typically require extensive labeled datasets or self-supervised signals, and the training process is time-consuming. Furthermore, domain adaptation issues may arise when applying the trained models to new scenarios—such as different urban structures or LiDARs with varying numbers of beams \cite{9925083, ding2024evolutionary}.


\subsection{Multi-LiDAR Calibration Methods for Urban Intersection Scenarios} 
Urban intersections often face unstable GPS signals, building occlusions, and large numbers of traffic participants with complex motion patterns. Multi-LiDAR calibration methods that rely on extrinsic priors or GPS positioning \cite{song2024spatial,VIeyeACM2021,VIPSACM2022,lu2023robust,xu2022v2x,hu2022where2comm} are prone to failure in such settings. Meanwhile, classical ICP or feature-based registration methods can perform poorly due to limited overlap and excessive noise. Although deep learning algorithms offer potential for tackling complex environments, they still face challenges related to data collection, model generalization, and computational efficiency in real-world applications. Additionally, when calibrating LiDARs with different configurations (such as beam counts, fields of view, resolutions, or scan frequencies), discrepancies in data resolution and feature representation must be carefully addressed \cite{huang2020feature, 9925083, ding2024evolutionary}.

In summary, multi-LiDAR calibration for urban intersections remains an open problem, requiring a balance among accuracy, robustness, and real-time performance.

\begin{figure}[tbp] 
\centering 
\includegraphics[width=0.5\textwidth]{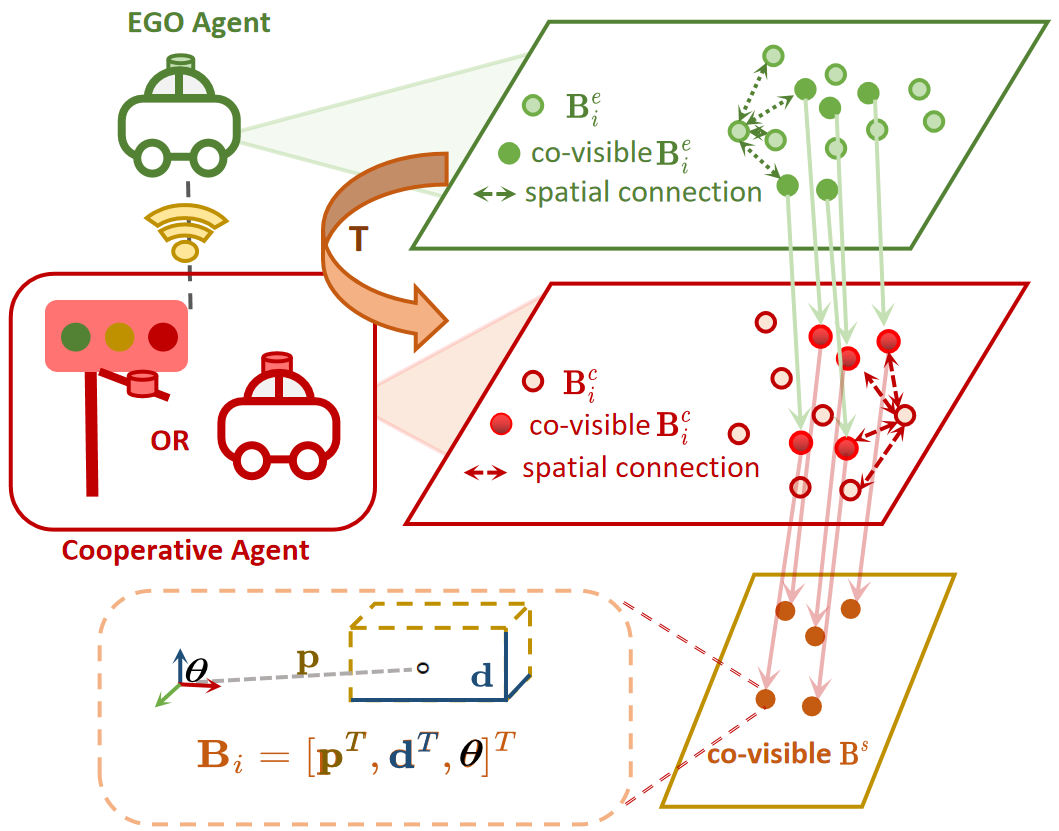} 
\caption{Diagram of notations in problem formulation.} 
\label{fig:basic_notions} 
\end{figure}




\section{PROBLEM FORMULATION}
\label{section:problem_formulation}

In this study, we construct a network of agents consisting of an Ego Agent and a Cooperative Agent (e.g., connected vehicles or roadside units), denoted as \( \mathbf{A} = \{e, c\} \). We refer to the objects detected by these agents as passive objects, which do not participate in the information exchange. These objects serve as scene features, enriching the environmental description. Each object in the set of passive objects \( \mathbf{B} \) perceived in the environment is represented as \( \mathbf{B}_{i} = [\mathbf{p}^T, \mathbf{d}^T, \theta]^T \in \mathbb{R}^7 \), where \( \mathbf{p} \in \mathbb{R}^3 \) denotes the center position, \( \theta \in [0, 2\pi) \) represents the orientation, and \( \mathbf{d} \in \mathbb{R}^3 \) represents the 3D dimensions. We can also express \( \mathbf{B}_i \) as a vertex matrix \( \hat{\mathbf{B}}_i \). Here, \( \hat{\mathbf{B}}_i \in \mathbb{R}^{3 \times 8} \) represents the eight-vertex bounding box of the passive object, where each column corresponds to a 3D coordinate of a vertex. For convenience, the two representations \( \mathbf{B}_i \) (emphasizing object aggregation) and \( \hat{\mathbf{B}}_i \) (highlighting matrix-based transformations) will be used interchangeably to describe these passive objects in subsequent contexts.



From the perspectives of the Ego Agent and Cooperative Agent, the perceived object sets \( \mathbf{B}^{\text{e}} \) and \( \mathbf{B}^{\text{c}} \) are obtained, and their shared perceived objects are represented as \( \mathbf{B}^{\text{s}} = \mathbf{B}^{\text{e}} \cap \mathbf{B}^{\text{c}} \).

It is important to note that \( \mathbf{B}^{\text{e}} \) and \( \mathbf{B}^{\text{c}} \) are located in their respective sensor coordinate systems, denoted as \( {}^{E}\mathbf{B}^{\text{e}} \) and \( {}^{C}\mathbf{B}^{\text{c}} \). However, for simplicity, we only label cross-coordinate sets, such as \( {}^{C}\mathbf{B}^{\text{e}} \). Elements in \( \mathbf{B}^{\text{s}} \) have been transformed into a unified coordinate system and will not be explicitly discussed further.


Previous studies typically rely on positioning systems to obtain a prior extrinsic, primarily focusing on solving the extrinsic optimization problem. In this work, we aim to eliminate the need for prior extrinsic by fully exploiting relationships between \(\mathbf{B}^{\text{e}}\) and \(\mathbf{B}^{\text{c}}\).

Then, a feature point cloud \( \mathbf{P} \) is constructed from the object set \( \mathbf{B}^{\text{s}} \) as follows:
\begin{equation}
\label{algorithm:feature_pointcloud}
\mathbf{P} = \left[ \hat{\mathbf{B}}^{\text{s}}_1 , \hat{\mathbf{B}}^{\text{s}}_2 , \cdots , \hat{\mathbf{B}}^{\text{s}}_n \right]^\top, \quad \mathbf{P} \in \mathbb{R}^{8n \times 3}
\end{equation}
where \(\hat{\mathbf{B}}^{\text{s}}_i\) represents the i-th target in \( \hat{\mathbf{B}}^{\text{s}} \), $n$ denotes the number of co-viewing boxes.

The ultimate goal is to obtain the optimal extrinsic parameters \( \hat{\mathbf{T}} \) by minimizing the distance between the corresponding feature point clouds:
\begin{equation}
\label{eq:Rt}
(\hat{\mathbf{R}}, \hat{\mathbf{t}}) = \arg\min_{\mathbf{R}, \mathbf{t}} \mathbf{E}(\mathbf{P}^{\text{e}}, \mathbf{R} \mathbf{P}^{\text{c}} + \mathbf{t})
\end{equation}
\begin{equation}
\label{eq:T}
\hat{\mathbf{T}} = \begin{bmatrix}
\hat{\mathbf{R}} & \hat{\mathbf{t}} \\
\mathbf{0}^T & 1
\end{bmatrix}
\end{equation}
where \( \hat{\mathbf{R}} \in SO(3) \) is the rotation matrix, and \( \hat{\mathbf{t}} \in \mathbb{R}^3 \) is the translation vector. Together, they form the extrinsic parameters \( \hat{\mathbf{T}} \in SE(3) \) as shown in Eq.(\ref{eq:Rt}). Eq.(\ref{eq:T}) performs the rotation and translation of the point cloud, transforming both point clouds into the same coordinate system. The point cloud error metric, \( \mathbf{E}(\cdot) \), is computed using the L2 norm (Euclidean distance) between the feature point clouds, as detailed in Eq.(\ref{eq:weighted_pointcloud_registration}).


\section{METHODOLOGY}
\label{section:methodology}

\subsection{Overview}
The proposed V2X-Calib++ framework addresses the multi-LiDAR calibration problem in V2X scenarios from a sensor perspective, while fundamentally solving a cross-view cross-source point cloud global registration problem at the data level. 
Instead of operating on dense, potentially noisy point clouds, which is computationally intensive and sensitive to outliers especially without initial alignment, our method operates at the semantic level of 3D detected objects. This stems from an intuitive hypothesis: proper alignment of multi-view co-visible objects implies valid extrinsic transformations between perspectives. This is particularly relevant for V2X urban intersections, which typically provide sufficient co-visible traffic participants (objects) to constrain the problem. Thus, we recast global point cloud registration primarily as a robust co-visible object matching problem, addressed in two main stages.

First, shared targets across agents are associated through spatial graph representations, as illustrated in Fig.~\ref{fig4:matching_transformation}, where nodes denote objects and edges encode relative geometric constraints. This graph-based association mechanism is detailed in Section~\ref{sec:multiend_association}. 

Subsequently, we convert matched targets into weighted feature point clouds \( \mathbf{P}^e \) and \( \mathbf{P}^c \) (Eq.~\ref{algorithm:feature_pointcloud}), where confidence scores \( \mathbf{M}_{i,j} \) guide a robust weighted SVD algorithm (Fig.~\ref{fig:comprison_wsvd_svd}) to solve extrinsic parameters, as elaborated in Section~\ref{sec:extrinsic_solution}. 

This dual-phase approach ensures calibration accuracy by prioritizing high-confidence matches while adaptively suppressing noisy detections through graph-derived geometric constraints, achieving real-time performance without prior pose initialization.

\subsection{Multi-End Object Association}
\label{sec:multiend_association}

\begin{figure}[tbp] 
\centering 
\includegraphics[width=0.5\textwidth]{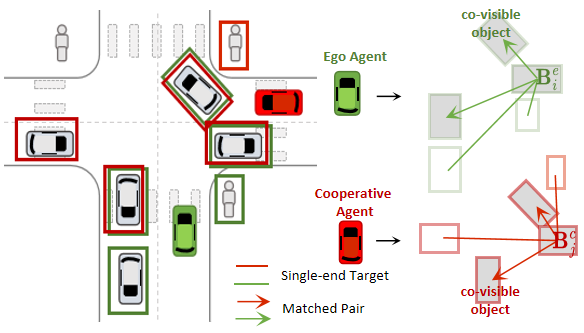} 
\caption{The core idea of object association is to transform the correspondence between targets into a correspondence of the spatial graph formed by the objects.} 
\label{fig4:matching_transformation} 
\end{figure}

Unlike existing object association methods that predominantly focus on short-term temporal continuity in object tracking scenarios\cite{wu2021deep}, our approach emphasizes scene-level co-visible object matching under large perspective variations, as illustrated in Algorithm \ref{alg:multiend}. We propose a dual-stage combinatorial strategy integrating singular value decomposition (SVD) and optimal transport (OT), with two main part: 1) \textbf{Scene-level correspondence mapping}: Transform potential pairwise object matches (\(\mathbf{B}^e_i, \mathbf{B}^c_j\)) into coordinate-unified spatial graphs (Fig.~\ref{fig4:matching_transformation}) through coordinate alignment (Eq.~\ref{eq:tranform}); 
2) \textbf{Optimal transport-based matching}: Convert the spatial graph association problem into a first-order node transportation task, where the \emph{Overall Distance} metric (Eq.~\ref{eq:overall_distance} and Eq.~\ref{eq:overall_confidence}) quantifies transportation costs by jointly evaluating match quantity (\( {}_\tau\overline{C}^{e_i}_{c_j} \)) and precision (\( {}_\tau\overline{D}^{e_i}_{c_j} \)), enabling robust object correspondence through optimal transport theory.

\begin{algorithm}[htbp]
\caption{Multi-end Object Association}
\label{alg:multiend}
    \begin{algorithmic}[1]
    \State \textbf{Input:} Objects \( \mathbf{B}^{e} \) and \( \mathbf{B}^{c} \) from Ego and Cooperative Agents
    \State \textbf{Output:} Matched object pairs \( \mathbf{B}^{s} \)
    \State Initialize affinity matrix \( \mathbf{M} \)
    \For{\(\mathbf{B}^{e}_{i}\) in \( \mathbf{B}^{e} \)}
        \For{\(\mathbf{B}^{c}_{j}\) in \( \mathbf{B}^{c} \)}
            \State Calculate \( \mathbf{F}^{i}_{j}(\cdot) \) using \text{Eq.}(\ref{eq:tranform})
            \State \( \mathbf{B}^{c} \) in Ego Coordinate  \( {}^{E_{i,j}}\mathbf{B}^{c} = \mathbf{F}^{i}_{j}(\mathbf{B}^{c}) \)
            \State Calculate \({}_\tau\overline{D}^{e_i}_{c_j}\) and \({}_\tau\overline{C}^{e_i}_{c_j}\) using \text{Eq.}(\ref{eq:overall_distance}) and \text{Eq.}(\ref{eq:overall_confidence})
            \If{ \( {}_\tau\overline{D}^{e_i}_{c_j} < \tau_1 \) }
                \State Update \( \mathbf{M}_{i,j} = {}_\tau\overline{C}^{e_i}_{c_j} \) 
            \Else
                \State Update \( \mathbf{M}_{i,j} = 0 \)
            \EndIf
        \EndFor
    \EndFor
    \State Get assignment matrix \( \mathbf{X} \) by solving \text{Eq.}(\ref{eq:assigndef}) using \cite{crouse2016implementing}
    \State Extract matched pairs \( \mathbf{B}^{S} \) from \( \mathbf{B}^{e} \) and \( \mathbf{B}^{c} \) using \( \mathbf{X} \)
    \State Return \( \mathbf{B}^{S} \)
    \end{algorithmic}
\end{algorithm}

\subsubsection{Correspondence Mapping}
\label{sec:correspondenceMapping}

Define \( \mathbf{B}^{e}_{i} \) and \( \mathbf{B}^{c}_{j} \) as the \( i \)-th and \( j \)-th perception objects in \( \mathbf{B}^{\text{e}} \) and \( \mathbf{B}^{\text{c}} \), respectively.
The first step in our association pipeline is to generate hypothetical alignments. For every potential pair of objects \( (\mathbf{B}^{e}_{i}, \mathbf{B}^{c}_{j}) \), we compute a relative transformation \( \mathbf{F}^{i}_{j}(\cdot) \) :
\begin{equation}
\label{eq:tranform}
    \mathbf{F}^{i}_{j}(\cdot) = \mathbf{R}^{i}_{j} \cdot + \mathbf{p}^{e}_{i} - \mathbf{R}^{i}_{j} \mathbf{p}^{c}_{j}
\end{equation}
where $\cdot$ is a placeholder for any input point set, \(\mathbf{p}^{e}_{i}\) and \(\mathbf{p}^{c}_{j}\) are center position of \( \mathbf{B}^{\text{e}} \) and \( \mathbf{B}^{\text{c}} \) respectively. \( \mathbf{R}^{i}_{j} \) is calculated as:
\begin{equation}
\label{eq:R}
    \mathbf{R}^{i}_{j} = \mathbf{U} \text{diag}(1, 1, \det(\mathbf{U} \mathbf{V}^T)) \mathbf{V}^T
\end{equation}
\begin{equation}
\label{eq:svd:H}
    \mathbf{H} = \hat{\mathbf{B}}^{e}_{i} (\hat{\mathbf{B}}^{c}_{j})^T = \mathbf{U} \mathbf{\Sigma} \mathbf{V}^T 
\end{equation}
Here, \( \hat{\mathbf{B}}^{e}_{i} \) and \( \hat{\mathbf{B}}^{c}_{j} \) are the vertex matrix representations of the objects \( \mathbf{B}^{e}_{i} \) and \( \mathbf{B}^{c}_{j} \), respectively, as defined in Section \ref{section:problem_formulation}. The matrices \( \mathbf{U} \) and \( \mathbf{V} \) are the left and right singular matrices obtained by applying Singular Value Decomposition (SVD) to \(\mathbf{H}\), following the method in \cite{SVDPAMI1987}.

Crucially, this initial SVD is not intended to yield the final extrinsic parameters. Instead, its purpose is to provide a localized transformation hypothesis. This hypothesis allows us to globally align the entire Cooperative Agent's perceived scene \( \mathbf{B}^{c} \) with the Ego Agent's scene \( \mathbf{B}^{e} \) if this specific object pair  \( (\mathbf{B}^{e}_{i}, \mathbf{B}^{c}_{j}) \) were a correct match. This potential localized alignment is fundamental for the subsequent Spatial Similarity Assessment.

\subsubsection{Spatial Similarity Assessment}
\label{sec:odist}

With the scene locally aligned based on a candidate object pair  \( (\mathbf{B}^{e}_{i}, \mathbf{B}^{c}_{j}) \) using the transformation \(\mathbf{F}^{i}_{j}(\cdot)\) (defined in Eq.\ref{eq:tranform}), we then assess the quality of this hypothetical alignment using our novel \emph{Overall Distance} (\emph{oDist}) metric (\emph{oDist}). This metric is a cornerstone of our association strategy, designed to quantify the scene-level spatial consistency that arises if the candidate pair  \( (\mathbf{B}^{e}_{i}, \mathbf{B}^{c}_{j}) \)  were indeed a correct match.

To compute oDist, let \( {}^{E_{i,j}}\mathbf{B}^{c} = \mathbf{F}^{i}_{j}(\mathbf{B}^{c}) \) denote the transformation of all objects \( \mathbf{B}^{c} \) from the Cooperative Agent to the Ego Agent's coordinate system, based on the alignment hypothesis derived from the specific pair \( (\mathbf{B}^{e}_{i}, \mathbf{B}^{c}_{j}) \). Within this hypothetically aligned scene, we define a valid matching pair set \( {}_\tau\mathcal{V}^{e_i}_{c_j} \). This set comprises pairs of objects—one from \( \mathbf{B}^{e} \) and one from the transformed \( {}^{E{i,j}}\mathbf{B}^{c} \)—that satisfy a distance threshold \( \tau \). The specific mathematical definition is given as:

\begin{equation}
\label{eq:valid_matching_set}
    {}_\tau\mathcal{V}^{e_i}_{c_j} = \{ (\mathbf{B}^{e}_{i'}, {}^{E_{i,j}}\mathbf{B}^{c}_{j'}) \mid  d(\mathbf{B}^{e}_{i'}, {}^{E_{i,j}}\mathbf{B}^{c}_{j'}) \leq \tau \} 
\end{equation}
\begin{equation}
\label{eq:valid_match}
    d(\mathbf{B}^{e}_{i'}, {}^{E_{i,j}}\mathbf{B}^{c}_{j'}) =  \alpha \lVert \mathbf{p}^{e}_{i'} - {}^{E_{i,j}}\mathbf{p}^{c}_{j'} \rVert + \beta \lVert \hat{\mathbf{B}}^{e}_{i'} - {}^{E_{i,j}}\hat{\mathbf{B}}^{c}_{j'} \rVert
\end{equation}

Here, \( \tau \) is a threshold designed to encompass potential matches, which can be empirically adjusted between 0 and 3 depending on the level of noise present in the scene. \( \mathbf{p}^{e}_{i'} \) represents the spatial center of the object \( \mathbf{B}^{e}_{i'} \), and \( \hat{\mathbf{B}}^{e}_{i'} \) represents the vertex matrix of the \(i'\)-th perception object. \( \alpha \) and \( \beta \) are weight factors that adjust the contribution of the location center and vertex distance of the detection frame to the index.

The \emph{oDist} itself consists of two key components, both calculated from the valid matching set \( {}_\tau\mathcal{V}^{e_i}_{c_j} \):





A confidence metric, \( {}_\tau\overline{C}^{e_i}_{c_j} \), which measures how many other object pairs become well-aligned (i.e., fall within \( {}_\tau\mathcal{V}^{e_i}_{c_j} \)) under the current transformation hypothesis. This directly indicates the degree of matching for the candidate pair \( (\mathbf{B}^{e}_{i}, \mathbf{B}^{c}_{j}) \). 

\begin{equation}
\label{eq:overall_distance}
    {}_\tau\overline{C}^{e_i}_{c_j} \dot{=} card({}_\tau\mathcal{V}^{e_i}_{c_j})
\end{equation}

A distance metric, \( {}_\tau\overline{D}^{e_i}_{c_j} \), which quantifies how close the geometric correspondence is for these well-aligned pairs by calculating their average distance. This provides an indication of the potential matching error for the candidate pair \( (\mathbf{B}^{e}_{i}, \mathbf{B}^{c}_{j}) \). 

\begin{equation}
\label{eq:overall_confidence}
    {}_\tau\overline{D}^{e_i}_{c_j} \dot{=} \frac{{}\sum_{(\mathbf{B}_{m} , \mathbf{B}_{n}) \in {}_\tau\mathcal{V}^{e_i}_{c_j} } d(\mathbf{B}_{m}, \mathbf{B}_{n})}{card({}_\tau\mathcal{V}^{e_i}_{c_j})}
\end{equation}

In these equations, \( \text{card}(\cdot) \) denotes the number of elements in a set, and \(d(\cdot, \cdot)\) is defined in Eq.(\ref{eq:valid_match}).

Therefore, the oDist, through its confidence \( {}_\tau\overline{C}^{e_i}_{c_j} \) and distance \( {}_\tau\overline{D}^{e_i}_{c_j} \) components, provides a rich, context-aware score. This score reflects the global plausibility of the initial local match \( (\mathbf{B}^{e}_{i}, \mathbf{B}^{c}_{j}) \) by considering its impact on the entire scene's consistency, moving significantly beyond simple pairwise object similarity. This comprehensive assessment is crucial for robustly identifying true correspondences in the subsequent object matching stage.

\subsubsection{Object Matching}

The robust oDist scores form the basis of our object matching stage. Then, we construct an affinity matrix \( \mathbf{M} \) between the perceived objects from the Ego Agent \( \mathbf{B}^{e} \) and the Cooperative Agent \( \mathbf{B}^{c} \). The elements \( \mathbf{M}_{i,j} \) of this matrix quantify the likelihood of \( \mathbf{B}^{e}{i} \) and \( \mathbf{B}^{c}_{j} \) being a correct match, based on the \emph{oDist} components:
\begin{equation}
    \mathbf{M}_{i,j} = \begin{cases} {}_\tau\overline{C}^{e_i}_{c_j} & \text{if } {}_\tau\overline{D}^{e_i}_{c_j} < \tau_1 , \\ 0 & \text{otherwise.} \end{cases}
\end{equation}
where \( \tau_1 \) represents a derived secondary filtering threshold, empirically adjustable between 0 and 2 based on the scene's noise level. This threshold applies to \( {}_\tau\overline{D}^{e_i}_{c_j} \), which is the average distance of all valid matching pairs in \( {}_\tau\mathcal{V}^{e_i}_{c_j} \). This is subtly different from \( \tau \) in Eq.(\ref{eq:valid_matching_set}), which refers to the distance between single potential matching pairs. The threshold \( \tau \) has a slightly larger range to include as many potential matching pairs as possible, whereas \( \tau_1 \) is used to minimize the impact of local spatial graph similarities on the calculation of valid matching pair similarity. This affinity matrix \( \mathbf{M} \) effectively captures the scene-level consistency for each potential object pair, as evaluated by oDist.

On this basis, we formulate an Optimal Transport (OT) problem to minimize the transformation cost from \( \mathbf{B}^{c} \) to \( \mathbf{B}^{e} \) based on \emph{oDist}. Considering constraints in real scenarios, we set two restrictions: 1) Each target in \( \mathbf{B}^{c} \) can be matched with at most one counterpart in \( \mathbf{B}^{e} \); 2) Not all targets can find matches due to differences in perception conditions such as field of view and occlusions.

Based on the above, we define the shared target matching task as follows:
\begin{equation}
\label{eq:assigndef}
\begin{aligned}
\underset{\mathbf{X}}{\text{argmin}} & \sum_{i=1}^{n} \sum_{j=1}^{m} - \mathbf{X}_{i,j} \mathbf{M}_{i,j} \qquad \text{s.t.} \quad \mathbf{X} \in \mathbf{\Pi} \\
\mathbf{\Pi} = \{ \mathbf{X} \mid  \mathbf{X} \in \{0, & 1\}^{n_1 \times n_2}, \, \mathbf{X} \mathbf{1}_{n_2} \leq \mathbf{1}_{n_1}, \, \mathbf{X}^T \mathbf{1}_{n_1} \leq \mathbf{1}_{n_2} \}
\end{aligned}
\end{equation}
where \( n \) and \( m \) represent the number of elements in \( \mathbf{B}^{e} \) and \( \mathbf{B}^{c} \) respectively, and \( \mathbf{X} \) is the assignment matrix. If \( \mathbf{X}_{i,j} = 1 \), it indicates that \( \mathbf{B}^{e}_{i} \) and \( \mathbf{B}^{c}_{j} \) are a matching pair, with the matching confidence given by \( \mathbf{M}_{i,j} \).
This constitutes a typical linear programming task, solved using \cite{crouse2016implementing}. The final set of shared objects \( \mathbf{B}^s \) can be represented as:
\begin{equation}
\label{eq:covisible_B}
\mathbf{B}^s = \{ (\mathbf{B}^{e}_{i}, \mathbf{B}^{c}_{j}, \mathbf{M}_{i,j} ) \mid \mathbf{X}_{i,j} = 1,  i \in [1, n], j \in [1, m] \}
\end{equation}

\subsection{Extrinsic Parameter Estimation}
\label{sec:extrinsic_solution}

\begin{figure}[t]
	\centering
	\setcounter{subfigure}{0} 
 
	\subfigure[Uniform Point Cloud Registration]{
		\begin{minipage}[t]{\linewidth}
			\centering
			\includegraphics[width=\linewidth]{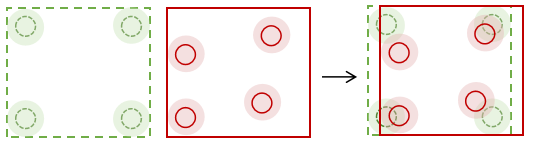}
			\label{fig:comprison_wsvd_svd:svd}
		\end{minipage}
	}
	\subfigure[Weighted Point Cloud Registration]{
		\begin{minipage}[t]{\linewidth}
			\centering
			\includegraphics[width=0.98\linewidth]{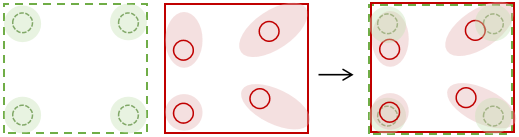}
			\label{fig:comprison_wsvd_svd:wsvd}
		\end{minipage}
	}
	\subfigure{
		\begin{minipage}[t]{\linewidth}
			\centering
			\includegraphics[width=0.8\linewidth]{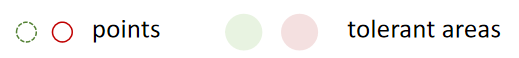}
		\end{minipage}
	}
	\caption{Comparative illustration of uniform and weighted point cloud registration. In weighted point cloud registration, points with lower confidence have a larger range of spatial tolerant areas, allowing points with higher confidence to predominantly influence the registration outcome.} 
	\label{fig:comprison_wsvd_svd} 
\end{figure}


Once the set of shared object pairs \( \mathbf{B}^s \) is established with associated matching confidences \( \mathbf{M}_{i,j} \) from the Optimal Transport (OT) stage Eq.(\ref{eq:covisible_B}), we proceed to estimate the final extrinsic parameters. The corners of these matched 3D detection boxes are used to form feature point clouds \( \mathbf{P}^e \) (for the Ego Agent) and \( \mathbf{P}^c \) (for the Cooperative Agent), as per Eq.(\ref{algorithm:feature_pointcloud}). Crucially, these are not uniform point clouds; each point pair derived from an object pair inherits the matching confidence \( \mathbf{M}_{i,j} \) established in the object association phase, effectively making \( \mathbf{P}^e \) and \( \mathbf{P}^c \) weighted feature point clouds. As illustrated in Fig. \ref{fig:comprison_wsvd_svd}, this weighting provides greater tolerance for points from less certain object matches, allowing the algorithm to prioritize the alignment of points from high-confidence object pairs. 

The weighted point cloud registration problem can be formulated as follows:

\begin{equation}
\label{eq:weighted_pointcloud_registration}
(\hat{\mathbf{R}}, \hat{\mathbf{t}}) = \arg\min_{\mathbf{R}, \mathbf{t}} \sum_{\mathbf{p}^e_i \in  \mathbf{P}^e , \mathbf{p}^c_i \in \mathbf{P}^c} w_i \| \mathbf{R} \mathbf{p}^e_i + \mathbf{t} - \mathbf{p}^c_i \|^2
\end{equation}
where $\mathbf{p}^e_i \in \mathbb{R}^3$ and $\mathbf{p}^c_i \in \mathbb{R}^3$ represent matched points in $\mathbf{P}^e$ and $\mathbf{P}^c$, respectively. The weight $w_{i}$ is the matching confidence \( \mathbf{M}_{i,j} \) from Eq.(\ref{eq:covisible_B}). $\hat{\mathbf{R}} \in SO(3)$ and $\hat{\mathbf{t}} \in \mathbb{R}^3$ represent the rotation matrix and translation vector of the target extrinsic parameters \( \hat{\mathbf{T}} \) in Eq.(\ref{eq:T}).

To solve this problem, we construct the weighted cross-covariance matrix as follows:

\begin{equation}
\overline{\mathbf{H}} = \sum_{i=1}^n w_{i} (\mathbf{p}^e_i - \overline{\mathbf{p}}_e)(\mathbf{p}^c_i - \overline{\mathbf{p}}_c)^T
\end{equation}
\begin{equation}
\overline{\mathbf{p}}_e = \frac{\sum_{i=1}^n w_i \mathbf{p}^e_i}{\sum_{i=1}^n w_i}, \quad \overline{\mathbf{p}}_c = \frac{\sum_{i=1}^m w_i \mathbf{p}^c_i}{\sum_{i=1}^m w_i}
\end{equation}
where $\overline{\mathbf{p}}_e$ and $\overline{\mathbf{p}}_c$ are the weighted centers of $\mathbf{P}^e$ and $\mathbf{P}^c$, respectively.

Subsequently, singular value decomposition (SVD) \cite{SVDPAMI1987} is applied to \( \overline{\mathbf{H}} \) to obtain the left singular matrix \( \mathbf{U} \) and right singular matrix \( \mathbf{V} \), and these are used in Eq.(\ref{eq:R}) to determine the rotation matrix \( \hat{\mathbf{R}} \). The translation vector $\hat{\mathbf{t}}$ can be obtained by:
\begin{equation}
    \hat{\mathbf{t}} = \mathbf{P}_e - \hat{\mathbf{R}} \mathbf{P}_c
\end{equation}

The final extrinsic parameters \( \hat{\mathbf{T}} \) can be obtained using Eq.(\ref{eq:T}).


\section{EXPERIMENT} 
\label{section:experiment}

In this chapter, we provide a detailed overview of the experimental design, evaluation metrics, dataset characteristics, and specific experimental setups used to assess the performance of our proposed V2I-Calib method.

\subsection{Evaluation Metrics}


To quantitatively evaluate our calibration method, we first define the fundamental error metrics for a single trial: the Relative Rotation Error \((\emph{RRE})\) and the Relative Translation Error \((\emph{RTE})\). Based on these, we establish three key performance indicators that will be used for our final analysis: the Success Rate at a given threshold \((SuccessRate@\lambda)\), the Mean Relative Rotation Error \((\emph{mRRE}@\lambda)\), and the Mean Relative Translation Error \((\emph{mRTE}@\lambda)\).

\textbf{Relative Rotation Error (\(\emph{RRE}\)):} Measures the accuracy of the rotational part of the calibration result, i.e., the angular difference between the estimated rotation matrix \( \mathbf{R}_e \) and the true rotation matrix \( \mathbf{R}_t \).

\begin{equation}
\emph{RRE} = \arccos\left(\frac{\text{tr}(\mathbf{\mathbf{R}_t^{-1}\mathbf{R}_e}) - 1}{2}\right) 
\end{equation}

\textbf{Relative Translation Error (\(\emph{RTE}\)):} Assesses the accuracy of the translation vector in the calibration result, i.e., the distance difference between the estimated translation vector \( \mathbf{t}_e \) and the true translation vector \( \mathbf{t}_t \).

\begin{equation}
\emph{RTE} = || \mathbf{t}_t^{-1} - \mathbf{t}_e ||_2
\end{equation}

These error metrics are termed 'relative' because they are computed against the ground-truth extrinsic parameters provided by the dataset. It is important to note the nature of this ground truth can differ. For simulated datasets, the ground truth is precise and absolute. For real-world datasets, however, the provided ``ground truth" is typically a high-quality estimate derived from post-processing and may exhibit minor inaccuracies in some scenarios.

\textbf{$SuccessRate@\lambda$:} The $SuccessRate@\lambda$ is defined as the proportion of calibration trials in a total sample set $ \mathcal{S} = \{1, 2, ..., N\} $ for which the achieved $ \emph{RTE} $ is below a predefined threshold $ \lambda $. For conciseness, this definition focuses solely on $ \emph{RTE} $, which is justified because, in relevant extrinsic calibration algorithms, $ \emph{RRE} $ and $ \emph{RTE} $ are often highly correlated, allowing one to serve as a general indicator of overall calibration quality—a practice also adopted in related literature \cite{HPCR_VIIV2023, VIPSACM2022, VIeyeACM2021}. This metric reflects the method's reliability in achieving a specified level of accuracy. The $SuccessRate@\lambda$ is mathematically expressed as: 
\begin{equation}
 \mathcal{S}^{\lambda}_{\textrm{valid}} = \{i\in\mathcal{S}|\emph{RTE}_i<\lambda\}
\end{equation}
\begin{equation}
\label{eq:success_rate} 
 SuccessRate@\lambda = \frac{ | \mathcal{S}^{\lambda}_{\textrm{valid}}|}{|\mathcal{S}|}
\end{equation}

While $SuccessRate@\lambda$ indicates how often the method succeeds, it does not quantify the average precision of these successful attempts. To provide this insight, we introduce \(\emph{mRRE}@\lambda\) and \(\emph{mRTE}@\lambda\). These metrics are calculated by averaging the \(\emph{RRE}\) and \(\emph{RTE}\) values exclusively over the subset of trials deemed successful by the threshold \(\lambda\). The metrics are then defined as:
\begin{equation}
\emph{mRRE}@\lambda = \frac{\sum_{i \in \mathcal{S}^{\lambda}_{\text{valid}}} \text{RRE}_i}{|\mathcal{S}^{\lambda}_{\text{valid}}|} 
\end{equation}
\begin{equation}
\emph{mRTE}@\lambda = \frac{\sum_{i \in \mathcal{S}^{\lambda}_{\text{valid}}} \text{RTE}_i}{|\mathcal{S}^{\lambda}_{\text{valid}}|} 
\end{equation}

These metrics offer a more robust assessment of the method's typical accuracy. This filtering is necessary because the outlier results from failed trials—where the (\emph{RTE}) can be excessively large (e.g., greater than 10 or even 100 meters)—are statistically meaningless for evaluating typical performance. Including them in a global average would heavily skew the results and obscure the method's true precision on valid alignments.

The choice of the threshold $ \lambda $ is critical and is guided by the operational requirements of Vehicle-to-Everything (V2X) applications, particularly for downstream tasks like cooperative perception and data fusion\cite{xu2022v2x, lu2023robust, hu2022where2comm,VIPSACM2022,HPCR_VIIV2023}. Drawing from existing literature and V2X system considerations (e.g., \cite{VIPSACM2022} suggests $ \lambda = 2\text{m} $, and \cite{HPCR_VIIV2023} suggests $ \lambda = 3\text{m} $), and with recommendations from some perception algorithms \cite{xu2022v2x, lu2023robust, hu2022where2comm}, an $ \emph{RTE} $ threshold $ \lambda $ in the range of 1 to 3 meters is often considered an acceptable upper bound for maintaining the requisite accuracy of these downstream tasks.

\subsection{Dataset}

In this study, we used the simulated dataset V2X-Sim \cite{li2022v2x} and the real-world dataset DAIR-V2X \cite{dairv2xCVPR2023} for experimental validation. Both datasets contain extensive data collected from Vehicle-Everything Cooperative Autonomous Driving (VXCAD) scenarios, including LiDAR data from vehicles and infrastructure as well as their 3D bounding box annotations and ground truth extrinsic parameters. The specifications of LiDAR Equipment are presented in Table \ref{tab:lidar_specs}. 

\begin{table}[htbp]
\centering
\caption{Specifications of LiDAR Equipment}
\label{tab:lidar_specs}
\renewcommand{\arraystretch}{1.2} 
\begin{tabular}{
  >{\centering\arraybackslash}m{0.45\columnwidth} 
  >{\centering\arraybackslash}m{0.07\columnwidth}
  >{\centering\arraybackslash}m{0.07\columnwidth}
  >{\centering\arraybackslash}m{0.07\columnwidth}
  >{\centering\arraybackslash}m{0.07\columnwidth}
}
\hline
\textbf{Parameter} & \multicolumn{2}{c}{\textbf{DAIR-V2X}} & \multicolumn{2}{c}{\textbf{V2X-Sim}} \\
\cline{2-3} \cline{4-5}
 & \textbf{R} & \textbf{V} & \textbf{R} & \textbf{V} \\
\hline
LiDAR Points (lines) & 300 & 40 & 32 & 32 \\
Horizontal Field of View (°) & 100 & 360 & 360 & 360 \\
Max Detection Range (m) & 280 & 200 & 70 & 70 \\
Volume (frames) & 3737 & 3737 & 1000 & 1000 \\
\hline
\end{tabular}
\medskip 
\vspace{0.5mm}
\footnotesize{\textit{Note: R = Roadside, V = Vehicle-side}}
\end{table}

One fundamental assumption for the effectiveness of the evaluation metrics discussed in the previous section is that the ground truth extrinsic parameters are sufficiently accurate. To justify this assumption, we analyze the processing workflows of the two datasets. V2X-Sim \cite{li2022v2x} utilizes the SUMO \cite{krajzewicz2012recent} and CARLA \cite{dosovitskiy2017carla} simulation platforms, which model the interaction between multiple vehicles and road-side units (RSUs) and the data acquisition process of their sensors. In this simulated environment, the data collection is considered to be free of delays and annotation errors, making the ground truth extrinsic parameters theoretically error-free. On the other hand, DAIR-V2X \cite{dairv2xCVPR2023} is based on real-world data collected from actual scenarios. After filtering, cleaning, localization system transformation, and pose refinement, relatively accurate ground truth extrinsic parameters are obtained. Although some errors still exist, they are generally considered to be within an acceptable range.

\subsection{Validation of Method Effectiveness}

In this section, we perform comprehensive experimental validations of our method using the datasets and metrics outlined above. This includes verifying the theoretical validity and noise sensitivity of the method on simulated datasets and evaluating its performance under real noise conditions on real-world datasets. All experiments were conducted on a device with an Intel i7-9750H CPU.

\subsubsection{Validity Verification on Simulated Dataset}
We first validate the theoretical feasibility of our method under these perfect conditions. The specific results are shown in Table \ref{tab:v2xsim_ideal_compare}.

\begin{table}[htbp]
\centering
\caption{Comparative Results on V2X-Sim.}
\begin{tabular}{@{}>{\centering\arraybackslash}p{2.0cm} >{\centering\arraybackslash}p{1.0cm} >{\centering\arraybackslash}p{1.2cm} >{\centering\arraybackslash}p{0.6cm} >{\centering\arraybackslash}p{0.6cm} >{\centering\arraybackslash}p{1.2cm}@{}}
\toprule
\multirow{2}{*}{\textbf{Method}} & $\emph{mRRE}$(\textdegree) & $\emph{mRTE}(m)$ & \multicolumn{2}{c}{$SuccessRate$(\%)} & \multirow{2}{*}{Time (s)} \\ 
\cmidrule(lr){2-2} \cmidrule(lr){3-3} \cmidrule(lr){4-5}
 & @3\textdegree & @3m & @1m & @2m & \\ 
\midrule
FGR\cite{zhou2016fast} & 0.69 & 0.16 & 78.64 & 95.15 & 0.92 \\
Quartro\cite{lim2022single} & 0.17 & 0.18 & 96.40 & 98.20 & 0.83 \\
Teaser++\cite{yang2020teaser} & 0.77 & 0.17 & 76.70 & 94.17 & 0.91 \\
V2I-Calib\cite{qu2024v2i} & 0.06 & 0.03 & 93.26 & 95.48 & 0.37 \\
\textbf{V2I-Calib++} & \textbf{0.01} & \textbf{0.01} & \textbf{96.80} & \textbf{98.31} & \textbf{0.13} \\
\bottomrule
\end{tabular}
\label{tab:v2xsim_ideal_compare}
\end{table}

We find that under conditions of perfect detection and perfect transmission, V2I-Calib++ achieves near-zero rotation and translation deviations quickly, proving the theoretical correctness of the perception-based approach. It is also observed that even in ideal conditions, the success rate of the method is not 100\%. This is due to the spatial information between perception objects occasionally having similarities, which can affect our method. However, these coincidental spatial similarities are rare, and the likelihood can be reduced by increasing the number of perception objects.

\subsubsection{Sensitivity to Noise}

To evaluate our method's robustness against realistic perception errors, we conducted a noise injection experiment on the V2X-Sim dataset \cite{li2022v2x}. We grounded our noise model in the performance of a widely-used 3D detector, PointPillars \cite{lang2019pointpillars}, on the nuScenes benchmark \cite{caesar2020nuscenes}. On this benchmark, PointPillars achieves a mean Average Translation Error (mATE) of $0.32m$ and a mean Average Orientation Error (mAOE) of $0.28 rad$ ($\approx 16^\circ$).

Recognizing that these are average metrics and that individual detections, especially in challenging scenarios, can exhibit larger deviations, our experiment was designed to span a wider spectrum. We injected Gaussian noise into the object's position (mean \( \mu_1 = 0 \, \text{m} \), with standard deviation from 0 to 2.0m) and von Mises noise into its orientation (mean \( \mu_2 = 0^\circ \), with standard deviation from 0° to 25°). This range effectively covers performance from typical error levels to more severe failure cases.

\begin{figure}[tp] 
\centering 
\includegraphics[width=0.5\textwidth]{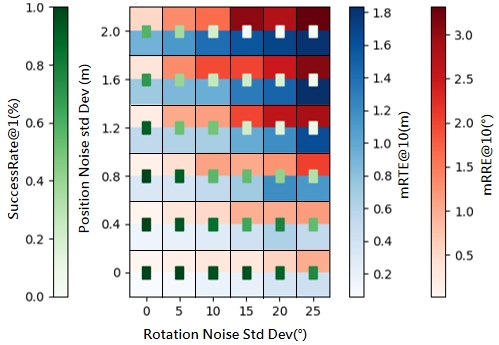} 
\caption{Heatmap Analysis of Three Performance Metrics Under Varying Noise Levels Applied to Ground Truth Bounding Boxes in the V2X-Sim Dataset. Here, we set the success rate standard to $\lambda=1m$. Despite increasing errors with higher input noise, the method shows strong noise resilience.} 
\label{fig:heatmap_v2xsim_noise_variation} 
\end{figure}

As shown in Fig.~\ref{fig:heatmap_v2xsim_noise_variation}, while the method is resilient to individual error dimensions, a combination of translation and rotation noise leads to a more rapid decline in effectiveness. Encouragingly, our method still demonstrates a strong corrective capability even under high noise. Specifically, the maximum \emph{mRTE} and \emph{mRRE} were controlled at $1.8m$ and $3.5^\circ$, respectively. This confirms the method's robustness across a realistic spectrum of perception noise.

\begin{table*}[htbp]
\centering
\begin{adjustbox}{max width=\textwidth}
\small 
\begin{threeparttable}
\caption{
Comparative Results on the DAIR-V2X Dataset. For the methods that require initial pose values, we add noise of equal magnitude to the rotational and translational dimensions to simulate different levels and sources of noise in real-world scenarios. Lower values are better for \emph{mRRE} and \emph{mRTE} ( $\downarrow$), and higher values are better for $SuccessRate$ ($\uparrow$). Subscripts GT, PP, and SC denote ground-truth boxes, PointPillars \cite{lang2019pointpillars} detector boxes, and SECOND \cite{yan2018second} detector boxes, respectively. The superscript k signifies the use of top-k dimension-sorted boxes, while \(\infty\) indicates use all boxes provided. The \textcolor{red}{\textbf{best}} and \textcolor{blue}{\underline{second-best}} results are highlighted in each section.}
\label{tab:contrast_experiment}
\begin{tabular}{@{}ccc*{3}{S[table-format=2.4]}*{3}{S[table-format=2.4]}*{3}{S[table-format=2.4]}c@{}}
\toprule
\multirow{2}{*}{Init} & 
\multirow{1}{*}{Noise} & 
\multirow{2}{*}{Method} & 
\multicolumn{3}{c}{\emph{mRRE} (\textdegree) $\downarrow$} & 
\multicolumn{3}{c}{\emph{mRTE} (m) $\downarrow$} & 
\multicolumn{3}{c}{$SuccessRate$ (\%) $\uparrow$} & 
\multirow{2}{*}{Time (s) $\downarrow$} \\
\cmidrule(lr){4-6} \cmidrule(lr){7-9} \cmidrule(lr){10-12}
 & \textbf{(m \& \si{\degree})} & & \textbf{@1\si{\degree}} & \textbf{@2\si{\degree}} & \textbf{@3\si{\degree}} & \textbf{@1m} & \textbf{@2m} & \textbf{@3m} & \textbf{@1m} & \textbf{@2m} & \textbf{@3m} & \\
\midrule
\multirow{12}{*}{\checkmark} 
 & 0 & \multirow{3}{*}{ICP \cite{besl1992method}} & 0.65 & 0.98 & 1.07 & \textcolor{red}{\textbf{0.42}} & \textcolor{red}{\textbf{0.54}} & \textcolor{red}{\textbf{0.58}} & 47.52 & 89.55 & 96.01 & 2.91 \\
 & 1 & & 0.80 & 1.36 & 1.72 & 0.66 & 1.31 & 1.62 & 0.86 & 37.93 
 & 80.50 & 2.92 \\
 & 2 & & 0.00 & 1.48 & 2.11 & 0.00 & 1.33 & 2.03 & 0.00 & 3.66 & 19.94 & 2.86 \\
\cmidrule(l){2-13}
 & 0 & \multirow{3}{*}{PICP \cite{serafin2017using}} & \textcolor{red}{\textbf{0.52}} & \textcolor{red}{\textbf{0.80}} & \textcolor{red}{\textbf{0.88}} & \textcolor{red}{\textbf{0.42}} & \textcolor{red}{\textbf{0.54}} & \textcolor{blue}{\underline{0.57}} & \textcolor{red}{\textbf{59.59}} & \textcolor{red}{\textbf{90.41}} & \textcolor{blue}{\underline{96.12}} & 1.35 \\
 & 1 & & 0.74 & 1.31 & 1.67 & 0.75 & 1.32 & 1.63 & 2.91 & 42.78 & 87.93 & 1.76 \\
 & 2 & & 0.80 & 1.40 & 2.11 & 0.53 & 1.45 & 2.10 & 0.22 & 2.69 & 21.12 & 1.70 \\
\cmidrule(l){2-13}
 & 0 & \multirow{3}{*}{VIPS \cite{VIPSACM2022}} & 0.63 & \textcolor{blue}{\underline{0.89}} & \textcolor{blue}{\underline{0.99}} & 0.54 & 0.78 & 0.89 & \textcolor{blue}{\underline{54.20}} & \textcolor{blue}{\underline{88.69}} & \textcolor{red}{\textbf{97.63}} & 0.46 \\
 & 1 & & 0.66 & 1.04 & 1.24 & 0.54 & 0.82 & 1.02 & 18.53 & 39.01 & 47.74 & \textcolor{blue}{\underline{0.44}} \\
 & 2 & & \textcolor{blue}{\underline{0.58}} & 1.17 & 1.56 & \textcolor{blue}{\underline{0.48}} & 0.96 & 1.39 & 2.37 & 7.87 & 13.15 & 0.47 \\
\cmidrule(l){2-13}
 & 0 & \multirow{3}{*}{CBM \cite{song2024spatial} $\dagger$ } & 0.61 & 0.97 & 1.21 & 0.53 & 0.80 & 1.06 & 17.11 & 23.04 & 26.49 & \textcolor{red}{\textbf{0.35}} \\
 & 1 & & 0.71 & 0.94 & 1.14 & 0.61 & \textcolor{blue}{\underline{0.74}} & 1.00 & 9.91 & 15.63 & 16.49 & 0.36 \\
 & 2 & & 0.69 & 1.09 & 1.38 & 0.58 & 0.76 & 1.06 & 6.03 & 12.28 & 16.81 & \textcolor{red}{\textbf{0.35}} \\
\midrule
\multicolumn{1}{c}{\multirow{8}{*}{\texttimes}}
 & - & FGR\cite{zhou2016fast} & 0.71 & 1.15 & 1.47 & 0.70 & 1.13 & 1.45 & 14.76 & 31.57 & 35.34 & 22.73 \\
 & - & Quartro\cite{lim2022single} & \textcolor{red}{\textbf{0.62}} & 1.22 & 1.46 & 0.65 & 1.19 & 1.51 & 12.07 & 30.50 & 45.04 & 21.58 \\
 & - & Teaser++\cite{yang2020teaser} & 0.69 & 1.13 & 1.47 & 0.66 & 1.09 & 1.44 & 14.33 & 29.74 & 34.81 & \textcolor{blue}{\underline{22.43}} \\
 \cmidrule(l){2-13} 
& - & V2I-Calib\cite{qu2024v2i} & 0.66 & \textcolor{blue}{\underline{1.03}} & \textcolor{blue}{\underline{1.25}} & 0.54 & 0.91 & 1.18 & 25.54 & 55.93 & 72.31 & 0.21 \\
 & - & V2I-Calib++\textsubscript{GT}\textsuperscript{\(\infty\)} & \textcolor{red}{\textbf{0.62}} & \textcolor{red}{\textbf{1.01}} & 1.26 & \textcolor{red}{\textbf{0.49}} & \textcolor{red}{\textbf{0.83}} & 1.07 & 22.88 & 48.03 & 61.49 & 0.46 \\
 & - & V2I-Calib++\textsubscript{GT}\textsuperscript{25} & \textcolor{blue}{\underline{0.63}} & \textcolor{red}{\textbf{1.01}} & \textcolor{red}{\textbf{1.23}} & \textcolor{blue}{\underline{0.52}} & \textcolor{blue}{\underline{0.85}} & \textcolor{red}{\textbf{1.05}} & \textcolor{red}{\textbf{32.27}} & \textcolor{red}{\textbf{67.59}} & \textcolor{red}{\textbf{82.93}} & 0.12 \\
 & - & V2I-Calib++\textsubscript{GT}\textsuperscript{15} & 0.65 & 1.05 & 1.30 & 0.54 & 0.87 & 1.10 & \textcolor{blue}{\underline{26.79}} & \textcolor{blue}{\underline{61.17}} & \textcolor{blue}{\underline{78.75}} & \textcolor{blue}{\underline{0.09}} \\
 & - & V2I-Calib++\textsubscript{GT}\textsuperscript{10} & 0.66 & 1.11 & 1.36 & 0.57 & 0.92 & 1.15 & 20.02 & 54.86 & 71.98 & \textcolor{red}{\textbf{0.04}} \\
 & - & V2I-Calib++\textsubscript{PP}\textsuperscript{15} & 0.66 & 1.06 & 1.29 & 0.55 & 0.86 & 1.07 & 24.91 & 56.62 & 70.94 & - \\
 & - & V2I-Calib++\textsubscript{SC}\textsuperscript{15} & 0.65 & 1.05 & 1.29 & 0.54 & 0.86 & \textcolor{blue}{\underline{1.06}} & 25.15 & 56.89 & 71.23 & - \\
\cmidrule(l){2-13}
 & - & V2I-Calib++\textsubscript{GT}\textsuperscript{25}(hSVD)$\ddagger$ & 0.71 & 1.13 & 1.35 & 0.62 & 0.98 & 1.25 & 21.82 & 60.43 & 74.92 & 0.12 \\
 & - & V2I-Calib++\textsubscript{GT}\textsuperscript{25}(mSVD)$\ddagger$ & 0.67 & 1.08 & 1.31 & 0.56 & 0.94 & 1.19 & 25.22 & 63.58 & 80.22 & 0.12 \\
\bottomrule
\end{tabular}
\begin{tablenotes}
\small
\item[$\dagger$]: For CBM \cite{song2024spatial}, our reimplementation (due to partial code availability) achieves comparable accuracy but significantly lower success rates under $ SuccessRate@\lambda $. We will make it open-source in our codebase.
\item[$\ddagger$]: V2I-Calib++ entries without parentheses (e.g., V2I-Calib++\textsubscript{GT}\textsuperscript{25}) use the proposed Weighted SVD (wSVD) by default. Comparisons between wSVD, mSVD, and hSVD strategies (Section~\ref{sec:ablation:extrinsic_strategy}) validate wSVD's superior robustness.
\end{tablenotes}
\end{threeparttable}
\end{adjustbox}
\end{table*}

\subsubsection{Validity Verification on Real-World Dataset}

\begin{figure*}[htbp]
	\centering
	\setcounter{subfigure}{0} 

    \subfigure[Ideal Scenario from V2X-Sim]{
		\begin{minipage}[t]{0.23\linewidth}
			\centering
			\includegraphics[width=\linewidth]{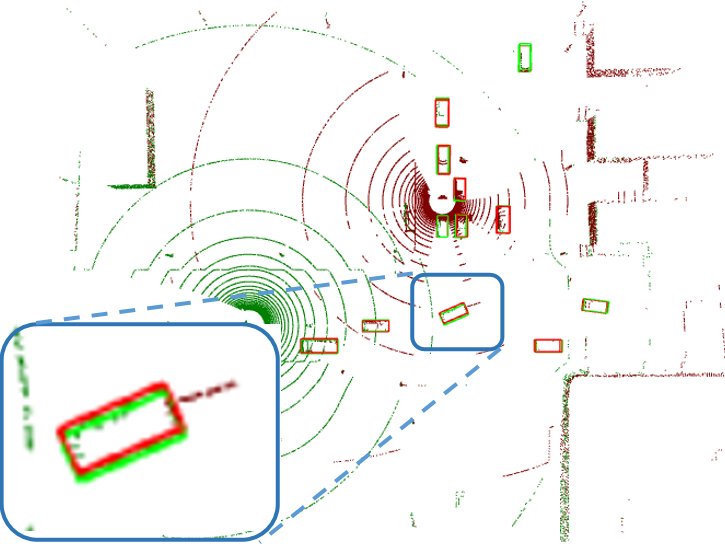}
			\label{fig:testresult:v2xsim}
		\end{minipage}
	}
	\subfigure[Ideal Scenario from DAIR-V2X]{
		\begin{minipage}[t]{0.23\linewidth}
			\centering
			\includegraphics[width=\linewidth]{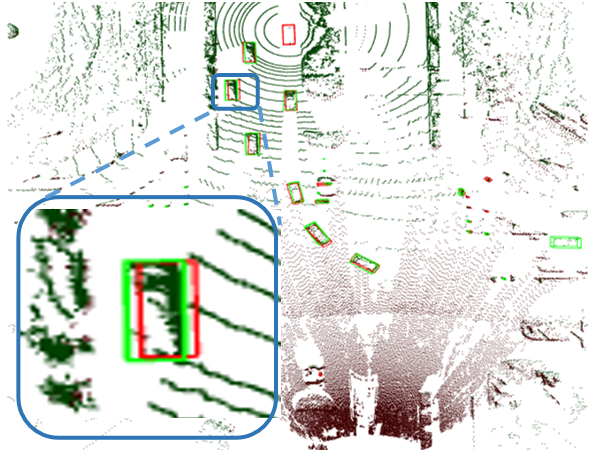}
			\label{fig:testresult:sub1}
		\end{minipage}
	}
	\subfigure[Challenging Scenario]{
		\begin{minipage}[t]{0.23\linewidth}
			\centering
			\includegraphics[width=\linewidth]{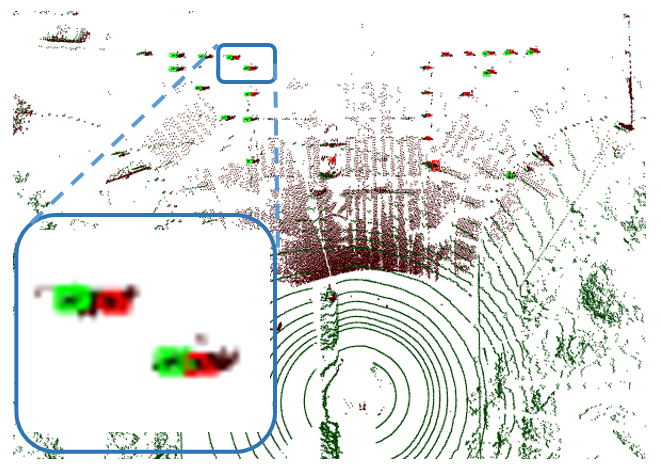}
			\label{fig:testresult:sub2}
		\end{minipage}
	}
	\subfigure[Complex Scenario]{
		\begin{minipage}[t]{0.23\linewidth}
			\centering
			\includegraphics[width=\linewidth]{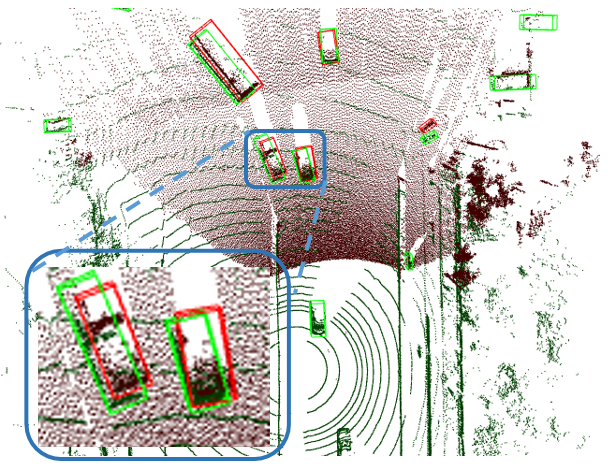}
			\label{fig:testresult:sub3}
		\end{minipage}
	}
	\caption{Comparative Calibration Results Across Diverse Scenarios: (a, d) Accurate alignment of multi-end perception objects. (b) Calibration errors due to small-sized perception objects. (d) Suboptimal calibration results due to inherent perception errors. } 
	\label{fig:testresult} 
\end{figure*}

To assess our method's practical efficacy, we performed experiments on the DAIR-V2X dataset \cite{dairv2xCVPR2023}. Its varied LiDAR configurations (Table~\ref{tab:lidar_specs}) and sparse point cloud overlaps (Fig.~\ref{fig:testresult}) pose substantial registration challenges.

\noindent\textbf{Performance and Accuracy.}

\textit{Comparison with methods requiring initial values:}
The upper part of Table~\ref{tab:contrast_experiment} contextualizes our results against methods reliant on initial poses. Notably, real-world V2I point clouds suffer from heterogeneity and cross-view discrepancies. Coupled with potential inaccuracies in DAIR-V2X ground-truth extrinsics, even initial-value-based methods \cite{ICPTPAMI1992, serafin2017using} under ideal conditions ($Noise = 0m \ \& \ 0^\circ$) exhibit an approximate 0.5m and \(0.5^\circ\) error. This can be seen as an accuracy \textit{upper bound} for this dataset. Introducing equivalent rotational and translational noise to these methods rapidly degrades their accuracy, rendering them nearly unusable at $2m \ \&\ 2 ^\circ$ noise. V2X-Calib++, however, effectively mitigates initial pose deviations. Its performance remains comparable to the dataset accuracy upper bound (e.g., the \(\emph{mRTE}@1m\)) and demonstrates stability against increasing initial errors, underscoring its practical value. The performance of V2X-Calib++ is primarily dictated by perceptual uncertainty and the number of detection boxes. For rigor, we will focus on its performance and key observations on DAIR-V2X rather than absolute numerical limits.

\textit{Comparison with initial-value-free methods:}
The lower part of Table~\ref{tab:contrast_experiment} shows V2X-Calib++ significantly surpassing other initial-value-free techniques. It achieves a high \(SuccessRate@2m\) of 56.89\% and low \(\emph{mRTE}@2m\) of 0.86m, even with real-world detector outputs from SECOND \cite{yan2018second} and PointPillars \cite{lang2019pointpillars}. This suggests that employing detection boxes as an intermediate representation for global registration of cross-source point clouds with large FoV differences can match the accuracy of point-based or feature-based methods. Crucially, it also enables real-time processing and substantially reduces transmission bandwidth, making it well-suited for V2X challenges.

An interesting observation is that while point/feature-based methods \cite{ICPTPAMI1992, serafin2017using} achieve higher accuracy with initial values than detection-box-based methods \cite{song2024spatial, VIPSACM2022} , while ours are more accurate without them, outperforming methods like \cite{yang2020teaser, lim2022single, zhou2016fast}.

\noindent\textbf{Impact of Detection Uncertainty.}
To quantify the impact of real-world perception noise, we used detections from PointPillars (PP) \cite{lang2019pointpillars} and SECOND (SC) \cite{yan2018second} instead of ground-truth (GT) boxes. While perception errors cause slight performance degradation, V2X-Calib++ remains robust overall. The impact of detector errors on overall calibration accuracy in real scenes is considerably smaller than in simulations (Fig.~\ref{fig:heatmap_v2xsim_noise_variation}), sometimes even yielding better \(\emph{mRTE}@3m\) values. The main adverse effect of detection errors is on \(SuccessRate@\lambda\). We posit this is due to inherent spatial deviations in V2I scenarios, which dilute the influence of perceptual errors on extrinsic parameters in successful calibrations. Stronger perceptual errors tend to cause registration failure and are thus excluded from accuracy statistics. The overall decrease in \(SuccessRate@\lambda\) is acceptable, reflecting our method's robustness to perceptual noise by emphasizing high-confidence matches. In this context, perceptual uncertainty's impact can be mitigated by pre-processing, such as expanding detection ranges and filtering low-confidence boxes, enhancing practical viability.

\noindent\textbf{Detection Box Quantity Effects.}
Our method's reliance on detection boxes means their quantity is a key factor, alongside perceptual uncertainty. DAIR-V2X frames exhibit wide variation in box counts (from a few to tens). Our SVD-based extrinsic search (Section~\ref{sec:correspondenceMapping}) naturally favors larger-dimension boxes due to greater spatial distinctiveness and potentially higher annotation/detection accuracy. We thus sorted boxes by volume and experimented with top-K subsets.
Table~\ref{tab:contrast_experiment} indicates that performance generally improves with more boxes. However, using all boxes (V2I-Calib++\textsubscript{GT}\textsuperscript{\(\infty\)}) degrades performance. Fig.~\ref{fig:testresult:sub2} illustrates this with small traffic cones from DAIR-V2X, which are challenging to annotate/detect accurately. Their low spatial distinctiveness leads to larger SVD-derived extrinsic errors, impacting overall performance. Thus, pre-filtering input boxes significantly boosts performance and reduces computation time.
Comparing V2I-Calib++\textsubscript{GT}\textsuperscript{15}, V2I-Calib++\textsubscript{PP}\textsuperscript{15}, and V2I-Calib++\textsubscript{GT}\textsuperscript{10} reveals that box quantity has a greater impact on the method's performance than individual box uncertainty. This aligns with our method's focus on information derived from the set of boxes, making set size (quantity) more influential than individual box characteristics (uncertainty).

\noindent\textbf{Runtime Analysis.}
Preliminary multi-threaded Python optimizations on an Intel i7-9750H CPU @2.6GHz yield a runtime of ~0.09s for V2I-Calib++\textsubscript{GT}\textsuperscript{15}, which is significantly below the $0.35s$ requirement in \cite{VIeyeACM2021}. This result is primarily intended for relative comparison with existing methods, highlighting the algorithm's real-time capability. Furthermore, the approach exhibits substantial parallelization potential. Specifically, the computational bottleneck arises from the $O(N^2)$ oDist calculations for potential pairs; however, these operations are mutually independent and thus amenable to parallelization. Additionally, the underlying matrix operations are well-suited for GPU acceleration. While the runtime upper bound occurs with V2I-Calib++\textsubscript{GT}\textsuperscript{\(\infty\)}(dozens of boxes), we note that such configurations may reduce accuracy, as discussed earlier. Therefore, practical deployment requires a trade-off between selected box quantity and runtime efficiency.

\subsection{Effectiveness Test of Overall Distance}

The multi-end calibration method proposed in this paper centers around the \emph{oDist} metric for achieving target association. Compared to the Overall Intersection over Union (oIoU) metric proposed in our previous method V2I-Calib \cite{qu2024v2i}, the \emph{oDist} exhibits superior performance and can better assess associations with objects that are further away. To validate the trends of the \emph{oDist} metric under various noise conditions, we designed experiments on DAIR-V2X \cite{dairv2xCVPR2023}.

\begin{figure*}[htbp]
	\centering
	\setcounter{subfigure}{0} 
	
	\subfigure[Comparison of metrics along the x-axis]{
		\begin{minipage}[t]{0.31\linewidth}
			\centering
			\includegraphics[width=\linewidth]{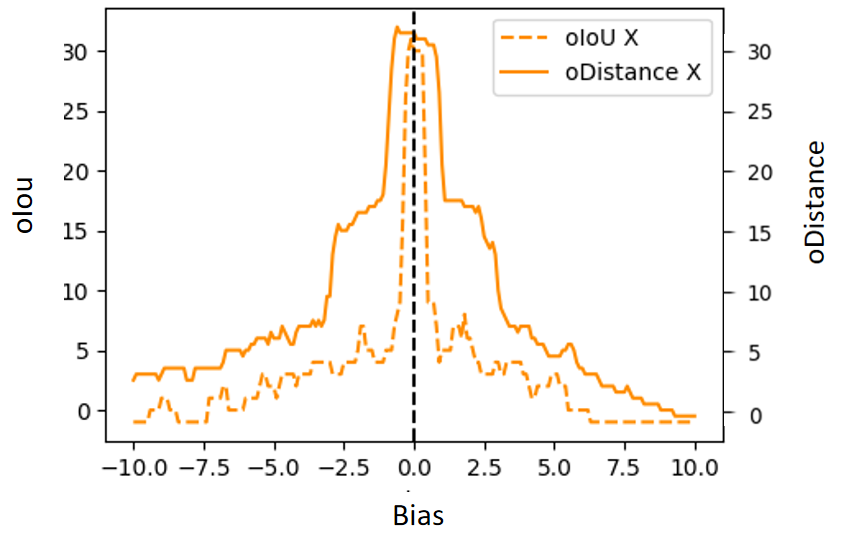}
			\label{fig:odistance_vs_oIoU:x}
		\end{minipage}
	}
	\subfigure[Comparison of metrics along the y-axis]{
		\begin{minipage}[t]{0.3\linewidth}
			\centering
			\includegraphics[width=\linewidth]{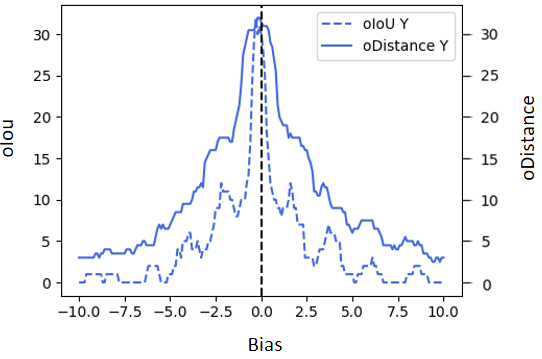}
			\label{fig:odistance_vs_oIoU:y}
		\end{minipage}
	}
	\subfigure[Comparison of metrics along the yaw dimension]{
		\begin{minipage}[t]{0.33\linewidth}
			\centering
			\includegraphics[width=\linewidth]{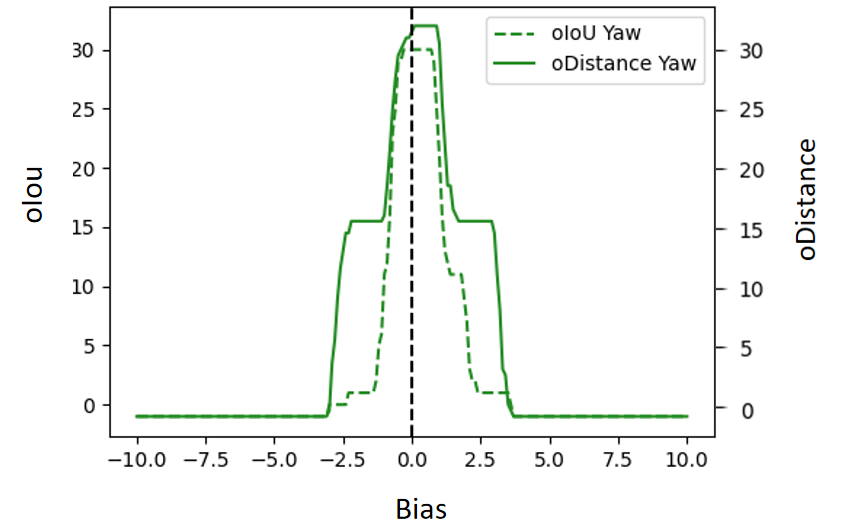}
			\label{fig:odistance_vs_oIoU:yaw}
		\end{minipage}
	}
	
	\caption{Verification of the indicator effects of the oDist metric proposed in this paper and the oIoU metric proposed in \cite{qu2024v2i} on the initial external parameters with added noise in different directions on DAIR-V2X. It is observed that the metric proposed by this method displays smoother curves, indicating better performance of oDist in monitoring the alignment level of external parameters. } 
	\label{fig:odistance_vs_oIoU} 
\end{figure*}

As shown in Fig. \ref{fig:odistance_vs_oIoU}, compared to the oIoU metric, the oDist proposed in this paper demonstrates a smoother curve of change with respect to different deviations in the external parameters. This implies that it is less susceptible to falling into local mathematical optima, thus providing better monitoring performance for external parameter alignment.

It is worth explaining that under continuously increasing biases, most perception-object-based metrics exhibit this type of local rise in values, due to the lack of confirmed associations between vehicle-road multi-end perception targets during the monitoring of external parameters. This causes an occasional rise in metrics when \( \mathbf{B}^e_i \) moves away from \( \mathbf{B}^c_j \) and closer to \( \mathbf{B}^c_{j+1} \).
In contrast to the oIoU metric, the oDist metric proposed in this paper extends the valid association phase from merely considering the IoU of two targets greater than zero to a set threshold distance determination, akin to the improvements suggested by softNMS\cite{bodla2017soft} on the classical NMS algorithm.

\subsection{Ablation Experiments}

In this section, we aim to validate the efficiency of the object association module introduced in Section \ref{sec:multiend_association} and the external parameter optimization module discussed in Section \ref{sec:extrinsic_solution} through a series of ablation experiments.

\subsubsection{Comparison of Object Association Strategies}

\begin{figure}[htbp] 
\centering 
\includegraphics[width=0.45\textwidth]{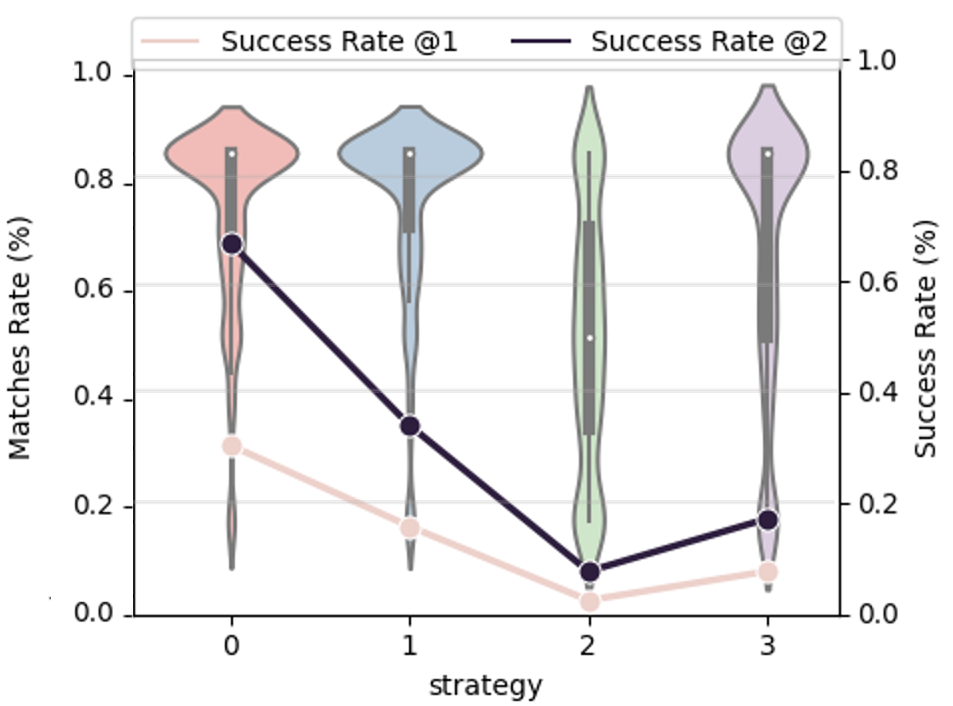} 
\caption{Violin plot comparing the effects of different object association strategies. The Matches Rate refers to the ratio of the number of correctly matched objects to the number of ground truth matched objects annotated in the scene. Strategy 1 is the oDist metric proposed in this paper, Strategy 2 is the oIoU metric proposed by \cite{qu2024v2i}, and Strategies 3 and 4 are the angular and length similarity metrics commonly used in papers such as \cite{VIPSACM2022}. The effectiveness of the target association strategy adopted in this paper is evident.} 
\label{fig:ablation_association} 
\end{figure}

To validate the superiority of the target association module strategy proposed in this paper, we conducted a comparative analysis with the angle-based and length-based association strategies defined in \cite{VIPSACM2022}, as well as the oIoU metric association strategy defined in \cite{qu2024v2i}. As shown in Fig. \ref{fig:ablation_association}, the proposed association strategy (Strategy 1) and the oIoU metric association strategy (Strategy 2) both achieved high target association rates. However, the oDist metric strategy excelled, achieving superior final success rates for both $\lambda=1$ and $\lambda=2$, thus confirming its robustness. The angle-based (Strategy 3) and length-based (Strategy 4) association strategies defined in \cite{VIPSACM2022}, originally intended for scenarios with initial external parameter values, did not perform as well overall. Nonetheless, they demonstrated some applicability, especially the length-based association strategy 4, which showed certain performance characteristics without initial external values.

These ablation studies not only confirm the effectiveness of each distinct module within our framework but also emphasize their collective role in enhancing the overall functionality of the system.

\subsubsection{Comparison of Strategies for External Parameter Solution}
\label{sec:ablation:extrinsic_strategy}
To validate the superiority of the external parameter optimization module discussed in Section \ref{sec:extrinsic_solution}, we compared three strategies: the Weighted SVD (wSVD) method proposed in this paper, the Mean SVD (mSVD) method, and the SVD method based on the highest confidence detection box (hSVD). As shown in Table. \ref{tab:contrast_experiment}, we observed that the Weighted SVD method proposed in this paper demonstrated enhanced robustness.



\section{Practical Use Cases}
\label{sec:practical_use_cases_v2x_calib_pp}

To demonstrate the practical value of V2X-Calib++, this section outlines a typical scenario illustrating its role in enhancing the robustness of V2X systems from initial deployment, particularly focusing on boot-up safety calibration and its synergy with positioning systems.



V2X-Calib++ can significantly enhance the safety and reliability of V2X systems during initial deployment and operation. It can operate independently without initial values (or  use these initial values as an reference) and leverage its online calibration capabilities for rapid optimization. Through the introduced $oDist$ metric, the system can continuously monitor the alignment quality of multi-end sensor data. If a calibration deviation is detected at system boot-up due to poor initial extrinsic parameters (e.g., a bad $oDist$ value), or if misalignment occurs during operation due to parameter drift, V2X-Calib++ will be triggered to quickly optimize the extrinsic parameters. This mechanism ensures that the system always operates on the basis of accurate calibration, achieving synergy with positioning systems: while the positioning system may degrade in challenging area like urban canyons, failing to provide adequate initial extrinsic parameters, V2X-Calib++ ensures precise, real-time sensor alignment at the perception level, which is crucial for immediate and ongoing V2X applications.

\begin{algorithm}[htbp]
\caption{V2X-Calib++ Calibration \& Monitoring}
\label{alg:bootup_monitoring_compact}
\begin{algorithmic}[1]
\State \textbf{Input:} 
\State \quad $P_1, P_2$: Detection boxes from two sensors
\State \quad $\theta_b$: Boot threshold for initial calibration
\State \quad $\theta_m$: Monitor threshold for runtime checks
\State \quad $R_{max}$: Max retry attempts
\State \textbf{Output:} $\mathbf{T}_{cur}$: Calibrated extrinsics
\State
\If{stored $\mathbf{T}_{stor}$ exists} 
    \State $\mathbf{T}_{cur} \gets \mathbf{T}_{stor}$ 
\Else 
    \State $\mathbf{T}_{cur} \gets \text{null}$ 
\EndIf
\State
\If{$\mathbf{T}_{cur} = \text{null}$ \textbf{or} $oDist(P_1,P_2,\mathbf{T}_{cur}) > \theta_b$}
    \State $\mathbf{T}_{cur} \gets \text{Calibrate}(P_1, P_2, \theta_b, R_{max})$
    \If{$\mathbf{T}_{cur} = \text{fail}$} \textbf{trigger alert} \EndIf
\EndIf
\State
\While{system running}
    \State Acquire new $P_1$, $P_2$
    \If{$oDist(P_1,P_2,\mathbf{T}_{cur}) > \theta_m$}
        \State $\mathbf{T}_{new} \gets \text{Calibrate}(P_1, P_2, \theta_m, R_{max})$
        \If{$\mathbf{T}_{new} \neq \text{fail}$} 
            \State $\mathbf{T}_{cur} \gets \mathbf{T}_{new}$; \textbf{store} $\mathbf{T}_{cur}$
        \Else
            \State \textbf{degrade operation}
        \EndIf
    \EndIf
    \State \textbf{apply} $\mathbf{T}_{cur}$
\EndWhile
\State
\Function{Calibrate}{$P_1, P_2, \theta, R$}
    \State $r \gets 0$
    \While{$r < R$}
        \State $\mathbf{T} \gets \text{V2X-Calib++}(P_1, P_2)$
        \If{$oDist(P_1,P_2,\mathbf{T}) \leq \theta$} 
            \State \Return $\mathbf{T}$ \Comment{Success}
        \EndIf
        \State $r \gets r+1$
    \EndWhile
    \State \Return \text{fail} \Comment{All attempts failed}
\EndFunction
\end{algorithmic}
\end{algorithm}

\section{DISCUSSION}
\label{section:discussion}

\subsection{Multi-End Calibration Tasks and Localization Tasks}
While localization and multi-end calibration are related, as both express spatial relationships, they are fundamentally distinct tasks in terms of their objectives, target objects, and role in the autonomous driving pipeline. Specifically, localization typically aims to determine a vehicle's absolute position within a global coordinate system, focusing on the vehicle's body frame. In contrast, calibration, the focus of this paper, seeks to resolve the relative pose transformation between different sensor coordinate systems. This places calibration as a foundational upstream task, responsible for enabling the accurate spatial alignment of sensor data, whereas localization is a downstream application that often relies on this pre-aligned data. Furthermore, the real-time nature of our method should not be misconstrued as a characteristic exclusive to localization. While classic calibration was often a static, offline process, modern targetless methods are increasingly evolving towards continuous, real-time operation to meet the demands of V2X systems. This trend positions real-time calibration as a crucial link between sensor alignment and perception, opening new avenues for end-to-end optimization. Therefore, based on its focus on relative sensor poses and its role as an upstream enabler for data fusion, our method firmly resides within the domain of sensor calibration.

\subsection{Trends in Multi-End Calibration}

The development of multi-sensor spatial calibration technology, as the basis for multi-sensor data fusion, largely depends on the demands of subsequent perception fusion tasks. Similarly, the trends in multi-end calibration tasks are influenced by multi-end perception fusion tasks. One trend in the latter is to achieve better perception effects under inaccurate external parameters \cite{xu2022v2x, wang2020v2vnet, vadivelu2021learning, yuan2022keypoints}. From another perspective, this integrates the calibration process into the perception task, no longer treating calibration as an independent output but as a dynamically adjusted intermediate quantity within the perception algorithm. This integration of calibration and perception aligns with the trend towards end-to-end development in autonomous driving. However, these methods still exhibit a low tolerance for deviations in external parameters. Under current research, multi-end external parameter calibration remains a necessary step.


\section{CONCLUSIONS}
\label{section:conclusions}

This paper addresses the challenges of complex traffic environments at urban intersections by introducing an innovative multi-end LiDAR calibration method—V2X-Calib++. The main contribution of V2X-Calib++ is that it overcomes the limitations of existing multi-end calibration methods that rely on positional priors, enabling effective calibration of multi-end LiDAR systems in environments with unstable positioning signals, such as urban canyons. The method combines a two-stage SVD algorithms and optimal transport theory to effectively solve the problem of data consistency among multi-end sensors. Additionally, the \emph{Overall Distance} metric (\emph{oDist}) proposed in this paper provides a reliable detection method for the degree of spatial alignment of scene external parameters.

Extensive experiments on the V2X-Sim and DAIR-V2X datasets validate the superiority of V2X-Calib++. It provides stable, high-precision calibration where initial-value-dependent methods fail due to noise, and it outperforms other initial-value-free approaches. The method is also robust to perception noise, effectively correcting errors from both simulated sources  and real-world detectors to maintain high performance. Its computational efficiency, with runtimes as low as 0.04 seconds, fully satisfies real-time demands. However, its primary limitation is a direct dependency on the quantity and quality of co-visible detection boxes. As performance can degrade with either too few objects or an excess of low-quality detections, this paper also proposes some pre-processing strategies in the experimental analysis section to mitigate these effects.

Future research will continue to explore the application potential of V2X-Calib++ in broader urban environments. This includes integrating other sensor types, leveraging static information from maps to extend its applicability to scenes with fewer dynamic objects, and scaling the algorithm for multi-agent systems. Expanding the current pairwise framework to robustly handle multiple agents in multi-frame asynchronous scenarios is a key direction, potentially by using our pairwise results as edges in a global pose-graph optimization network. However, this introduces communication bottlenecks, and future work will investigate co-optimizing calibration with communication strategies. On the performance front, given the algorithm's reliance on matrix operations, significant speed-ups are anticipated from GPU-accelerated implementations. Additionally, as autonomous driving technologies advance, integrating V2X-Calib++ more deeply with the full autonomous driving stack will be a critical area for subsequent research. We believe these advancements will not only promote the further development of Vehicle-to-Everything (V2X) technologies but also provide strong technical support for the intelligence and automation of future urban traffic systems.



\bibliographystyle{IEEEtranBST/IEEEtran}

\end{document}